# Enhancing Biomedical Relation Extraction with Directionality

Po-Ting Lai[1,†], Chih-Hsuan Wei[1,†], Shubo Tian[1], Robert Leaman[1] and Zhiyong Lu[1,*]

[1]Division of Intramural Research (DIR), National Library of Medicine (NLM), National Institutes of Health (NIH), Bethesda, MD 20894, USA

[†]The authors have contributed equally to this work.

[*]To whom correspondence should be addressed.

**Abstract**

Biological relation networks contain rich information for understanding the biological mechanisms behind the relationship of entities such as genes, proteins, diseases, and chemicals. The vast growth of biomedical literature poses significant challenges updating the network knowledge. The recent Biomedical Relation Extraction Dataset (BioRED) provides valuable manual annotations, facilitating the development of machine-learning and pre-trained language model approaches for automatically identifying novel document-level (inter-sentence context) relationships. Nonetheless, its annotations lack directionality (subject/object) for the entity roles, essential for studying complex biological networks. Herein we annotate the entity roles of the relationships in the BioRED corpus and subsequently propose a novel multi-task language model with soft-prompt learning to jointly identify the relationship, novel findings, and entity roles. Our results include an enriched BioRED corpus with 10,864 directionality annotations. Moreover, our proposed method outperforms existing large language models such as the state-of-the-art GPT-4 and Llama-3 on two benchmarking tasks. Our source code and dataset are available at https://github.com/ncbi-nlp/BioREDirect.
Contact: zhiyong.lu@nih.gov

## 1 Introduction

The exponential growth of biomedical literature raises a concern: How can researchers efficiently extract and synthesize crucial knowledge from vast data? Biomedical text mining methods provide researchers with a way to navigate knowledge. Biomedical Text Mining communities (Christopoulou, et al., 2019; Gurulingappa, et al., 2012; Herrero-Zazo, et al., 2013; Huang, et al., 2024; Kringelum, et al., 2016) are dedicated to advancing the field of Relation Extraction (RE). Much of this work has focused on RE within a limited snippet, such as single sentence (Miranda, et al., 2021; Peng, et al., 2024) or social media (Dirkson, et al., 2023). However, many statements of the biomedical relationships span multiple sentences (i.e., document-level). Xu, et al. (2016) reported that approximately 30% of chemical-induced disease relationships in the BC5CDR dataset (Wei, et al., 2016) cannot be found within a single sentence. Similarly, Islamaj, et al. (2024) revealed that a significant portion (27%) of RE false negative cases occur in inter-sentence contexts. These findings highlight the importance of studying document-level RE (Lai, et al., 2023; Luo, et al., 2022; Ming, et al., 2024; Wei, et al., 2016).

The recently created BioRED corpus (Luo, et al., 2022) encompasses document-level relationships across eight entity types: chemical-chemical, chemical-disease, chemical-gene, chemical-variant, disease-gene, disease-variant, gene-gene and variant-variant. Additionally, each relationship anno-tation is categorized by novelty to indicate whether the relationship represents a significant finding in the article or previously known background knowledge. BioRED corpus initially comprises 600 abstracts for RE system development and an additional 400 abstracts were annotated (Islamaj, et al., 2024) to enhance the coverage of emerging topics. Nonetheless, a key limitation remains in the BioRED corpus: it does not capture the directionality (subject/object) of the entity roles within the relationships. As shown in Fig. 1, while the figure indicates a positive correlation between the SAA gene and MAP kinases, it fails to specify the directionality of the relationship. As such, it becomes difficult to construct a meaningful biological network without clearly indicating the roles of the entities—for example, that SAA activates MAP kinases. This information needs to be included in downstream text mining research, such as studying the knowledge networks. To address this issue, in this work we first expand BioRED by adding the roles of the entities for the relationships in the red box highlighted in Fig. 1.

**Fig 1.** An example of relation directionality annotation in PMID:9746003 displayed on TeamTat

Recent biomedical RE approaches primarily utilized Bidirectional Encoder Representations from Transformers (BERT)-based classification. For instance, Luo, et al. (2022) developed two BERT-based models (Kenton and Toutanova, 2019) to classify relationships and novelty labels, respectively. They demonstrated simple but effective results of PubMedBERT (Gu, et al., 2021) over the BERT-graph transformer (BERT-GT) (Lai and Lu, 2021). Recently, Lai, et al. (2023) proposed BioREx, a data-centric approach for relation classification in BioRED. BioREx combines nine different biomedical relation extraction datasets and uses a hard (manually crafted) prompt to guide the BERT model for different tasks. This approach achieved high performance on BioRED corpus but focuses solely on relation type classification. Yuan, et al. (2024) combined a hierarchical tree graph (HTG) with a relation segmentation (RS) technique (HTGRS) to construct a bottom-up tree structure, with nodes representing context, mentions, sentences, and the overall document. These approaches often utilize BERT-based models with an input length limitation of 512 tokens. However, this limitation poses a critical challenge, as the textual evidence for some relationships in BioRED may appear at the end of a document, potentially exceeding BERT's input capacity. In response, our method herein incorporates a segmentation approach, similar to the method proposed by Huang, et al. (2022) for handling electronic health records (EHR), to address this issue and enhance overall performance.

Additionally, recent advances in pre-trained language models (PLMs) have shown that soft (virtual) prompts can outperform hard (manually crafted) prompts, learned embeddings that adapt to tasks during training, can outperform hard (manually crafted) prompts, pushing the boundaries of system performance. Lester, et al. (2021) demonstrated that soft prompts learned through backpropagation can be tuned to incorporate signals from any number of labeled examples. Peng, et al. (2024) further demonstrated the superiority of soft prompting over hard prompting in clinical concept recognition and end-to-end RE. Recent works have also demonstrated the capability of Large Language Models (LLMs) to handle various generation tasks. However, fine-tuning LLMs typically requires significant GPU resources due to the complexity of updating billions of parameters. The advent of parameter-efficient fine-tuning (PEFT) techniques (Dettmers, et al., 2024) offers a feasible solution. By adding adaptive parameters to LLMs and updating only these parameters, PEFT makes fine-tuning accessible to researchers with limited resources. Inspired by these technical advancements, we propose to integrate multi-task learning, soft-prompt learning, and a segmentation approach into BERT, and compare its performance with LLMs using PEFT on the BioRED and BC5CDR datasets. Overall, our contributions are summarized as follows:

- First, we annotate the entity roles (subject and object) and directionality for each relation pair on top of the BioRED relation annotations.
- We propose a joint multi-task model for extracting relations, novelties, and entity roles simultaneously.
- We introduce a segmentation approach that can effectively handle the input text with more than 512 tokens (the input limitation of the BERT-based models).
- We demonstrate that the use of soft-prompts dealing with the document-level RE can effectively improve the overall performance.
- We compare our method with state-of-the-art and fine-tuned large language models (LLMs) using PEFT techniques on two benchmark document-level RE tasks.

## 2 Materials and methods

In this section, we begin by outlining the definition and process of directional annotation for BioRED. Following that, we present the details of our proposed model, which integrates multi-task learning, context chunking, and soft-prompt techniques.

### 2.1 Annotating relation directionalities in BioRED

#### 2.1.1 Annotation Scope

Understanding the directionality of the relationships via the entity roles (subject and object) is fundamental for researchers aiming to uncover the mechanisms underlying complex biological processes, which is crucial for correctly interpreting the pathways. With information about entity roles, the questions "does a specific chemical inhibit the progression of a disease?" or "does it exacerbate the condition?" can be answered. Such insights are critical for developing effective therapeutic interventions. However, the entity roles of the relationhips in BioRED corpus were lacking. To enhance the utility of BioRED, we incorporated the entity role labels into the annotations, enabling researchers to capture the directionality of the relations.

We manually annotate the "subject" and "object" roles for the entities of each relation based on the context. For example, in "dexamethasone-induced hypertension", dexamethasone and hypertension are the subject and object, respectively. Comprehensive annotation guidelines, detailing the definitions and exception cases for "subject" and "object" across various entity pairs and relation types, can be found in the supplementary materials.

#### 2.1.2 Annotation Process

Building on the manually curated relation annotations in the original BioCreative VIII BioRED task (Islamaj, et al., 2023) dataset, which contains 1,000 abstracts, we added the subjects and objects of the entities in the relations. We employed TeamTat (Islamaj, et al., 2020), a powerful tool for annotating the entity roles of the relationships.

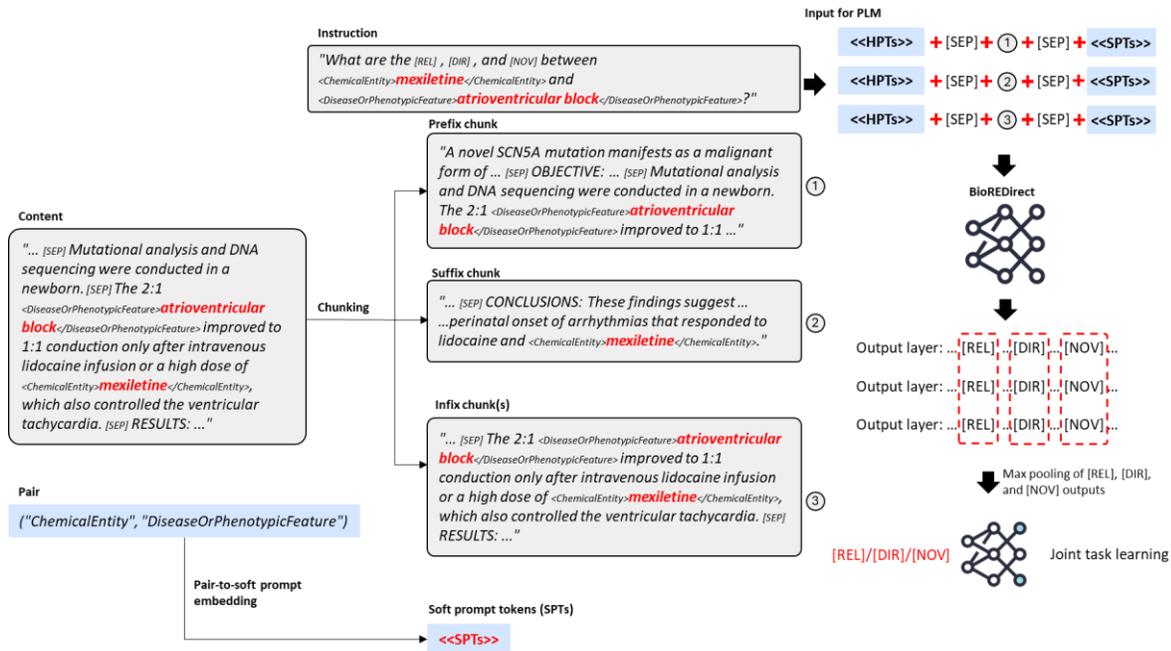

**Fig 2.** The overall architecture of our approach

Fig. 1 illustrates the TeamTat curation interface. It shows how each relationship is manually labeled on TeamTat. The subject of the pair within a relationship is manually designated with the "Subject" role, while the paired entity is assigned to the "Object" role.

We first randomly selected 50 abstracts for two annotators (with expertise in biomedical informatics) to practice. They followed the annotation guidelines and worked independently. After that, the two curators discussed the disagreements and summarized the rules needed for consistency. This phase is crucial for minimizing errors and ensuring accuracy. Any annotation discrepancies between the two annotators were discussed, leading to the refinements in our guidelines. Upon completing this initial annotation phase, we calculated the inter-annotator agreement (IAA) as the key metric of consistency for directionality annotations. With a high annotation consistency of 89.96% on the initial 50 articles, we proceeded to randomly assign the remaining 950 abstracts to the two annotators, with each abstract being annotated by one individual. Throughout this process, the annotators also documented all the unusual cases for later discussions and further summarized the agreement to the guidelines. The iterative feedback loops ensure quality of our annotations.

## 2.2 Relation Extraction

### 2.2.1 Problem Formulation

We defined this task as multi-label classification problem. Given a document $d$ containing a set of entities $E = \{e_i | i = 1, ..., n\}$, where $n$ is the number of entities. Our task is to classify each entity pair $\langle e_x, e_y \rangle$ into pre-defined label $y \in Y$, where $Y$ is a pre-defined label set of a task. Within document $d$, each entity $e_i$ may occur multiple times or be represented by synonyms. We break down the Relation Extraction (RE) task into three subtasks:

- Entity Pair Relation (REL): Identify and categorize the biomedical relationship of an entity pair $\langle e_x, e_y \rangle$. We use the original BioRED eight labels (e.g., positive correlation) with a "None" label for the pairs without a relationship. The set of the total nine relation classification labels is defined as $Y_{REL}$.
- Novelty (NOV): Determine whether the relationship represents a new scientific finding in the abstract. We labeled "Novel" for novel relation, "No" for background knowledge, and "None" for the pairs has no relation in $Y_{NOV}$.
- Directionality (DIR): Identify the roles of the two entities in the relationship. Four labels of $Y_{DIR}$ include: (1) Rightward: The first entity $e_x$ is the subject and initiates the relationship; (2) Leftward: The second entity $e_y$ is the subject and initiates the relationship; (3) Undirected: Directionality is unclear in the abstract. (4) None: No relation.

### 2.2.2 Input Text Representation

Fig. 2 illustrates the overall architecture of our proposed approach. We first employed SciSpacy (Neumann, et al., 2019) to split the document $d$ into sentences. Subsequently, BERT's tokenizer tokenizes the sentences into tokens. We concatenate the sentences using the special token [SEP].

Crucially, for each entity pair $\langle e_x, e_y \rangle$, boundary tags (e.g., <D> and </D> for diseases) are appended at the beginning and the end of the entities, where we used different tags for different entities. These tags are also added to the PLM's vocabulary to prevent them from being split into multiple tokens. As a result, a document $d$ with an entity pair $\langle e_x, e_y \rangle$, it can be represented as $t = [t_1, ..., t_m]$, where $t_i$ represents the $i^{th}$ token and $m$ is the length of the input sequence.

Because an input sequence $t$ can be longer than the maximum input length of common biomedical BERT models, we introduced a segmentation step to divide $t$ into three different chunks: prefix, suffix, and infix ($t_{\text{prefix}}$, $t_{\text{suffix}}$, and $t_{\text{infix}}$, respectively), which allows us to process longer documents without losing important context. Each chunk is a part of $t$. Most details of the chunks will be explained in the next section.

Additionally, for entity pair $\langle e_x, e_y \rangle$, we construct a hard prompt $p^{<x,y>} = [p_1^{<x,y>}, \ldots, p_k^{<x,y>}]$, acting as a guide to provide context about the entity pair and the specific RE task, as depicted in Fig. 2. The hard prompt contains three special tokens, [REL], [DIR], and [NOV], which are used in our classification module.

Furthermore, we employ a soft prompt $s$, consisting of a fixed length of learnable tokens, like special tokens that are learned during training, which allows the model to adapt to different tasks and domains. $s$ is defined as $s = [s_1, \ldots, s_q]$, where $q$ represents the length of the soft prompt. More results of the effect of the soft prompt length to the performance are in the Discussion section.

We concatenate the hard prompt <HPTs>, the three chunks, and the soft prompt with [SEP], which distinguish different components of the input. Consequently, the input embedding can be represented as follows:

$$H^{(0)} = [h_1^{(0)}, \ldots, h_L^{(0)}] = [\text{CLS}] \oplus p^{<x,y>} \oplus [\text{SEP}] \oplus t \oplus [\text{SEP}] \oplus s \quad (1)$$

where $h_i$ is the mapping of input $i^{\text{th}}$ token to trainable word embedding and $L$ is the total token length. $H^{(0)}$ will be fed into PLM, and $h_i^{(j)}$ denote the $i^{\text{th}}$ token's hidden vector in the $j^{\text{th}}$ layer of the PLM.

$$H^{(j)} = [h_1^{(j)}, \ldots, h_L^{(j)}] = \text{PLM}(H^{(0)}) \quad (2)$$

With three different chunks $t_{\text{prefix}}$, $t_{\text{suffix}}$, and $t_{\text{infix}}$, we represent them as $H_{\text{prefix}}^{(j)}$, $H_{\text{suffix}}^{(j)}$, and $H_{\text{infix}}^{(j)}$. This results in a context-aware representation for our multi-task learning.

### 2.2.3 Context Chunking

Due to potential information loss when the combined length of the input sequence $t$, hard prompt $p^{<x,y>}$, soft prompt $s$, and special tokens may exceed BERT's maximum sequence length, we implemented a chunking strategy. Specifically, we segmented $t$ into three overlapping chunks:

- Prefix chunk ($t_{\text{prefix}}$): The Prefix chunk comprises the initial $n$ tokens of the input sequence, where $n$ is calculated as the maximum number of tokens that, when combined with prompts and special tokens, can be accommodated within BERT's input maximum length.
- Suffix chunk ($t_{\text{suffix}}$): Similarly, the Suffix chunk consists of the final $n$ tokens of the input sequence, with $n$ determined using the same calculation method as the Prefix chunk to ensure maximum utilization of BERT's input capacity.
- Infix chunk ($t_{\text{infix}}$): For the Infix chunk, we identify the smallest window of sentences that encompass the entity pair, thereby focusing on the most relevant context. For instance, if three sentences contain the entity pair, the Infix chunk will comprise these three sentences, ensuring comprehensive coverage of the relevant information.

### 2.2.4 Classification Module

To effectively capture and utilize the contextual representation of specific tokens within a sequence, we designed a hidden token pooler module that enhances token representation by leveraging BERT's pre-trained pooler weights. These weights are the parameters in a linear transformation layer applied to the hidden state of the [CLS] token at the output of the BERT model. The pre-trained pooler is typically used when a single embedding representing the entire input sequence is needed, like classification task.

Our hidden token pooler module begins by initializing a fully connected layer with weights $W_j \in R^{d \times d}$ and biases $b_j \in R^{d \times 1}$ directly cloned from the BERT pooler's dense layer. $d$ is the dimension.

$$z_k = \tanh(W_k \cdot h_k^{(o)} + b_k) \quad (3)$$

where task $k \in \{\text{REL}, \text{DIR}, \text{NOV}\}$ and $o$ is the last hidden layer of PLM. $h_k^{(o)}$ is the hidden vector of the task token $t_k$ in the layer $o$. $z_k$ is fed into a classification module for the classification and as shown below:

$$\hat{y}_k = \tanh(W'_k \cdot z_k + b'_k) \quad (4)$$

where weights $W'_k \in R^{d_k \times d}$, biases $b'_k \in R^{d_k \times 1}$, and $d_k$ is the number of pre-defined classes of the task. We employ three distinct context-aware representations to generate corresponding output representations: $\hat{y}_{k,\text{prefix}}$, $\hat{y}_{k,\text{suffix}}$, and $\hat{y}_{k,\text{infix}}$. To aggregate the information from these output representations, we apply a max-pooling layer. This aggregation process refines the final output representation, denoted as $\hat{y}_k$, which is computed as follows:

$$\hat{y}_k = \max\{\hat{y}_{k,prefix}, \hat{y}_{k,suffix}, \hat{y}_{k,infix}\} \quad (5)$$

It allows for the integration of diverse contextual information, potentially enhancing the model's ability to capture relevant features across long contexts.

### 2.2.5 Objective Loss Function

We use binary cross-entropy with logits loss for three objectives: relation classification, novelty detection, and direction classification. This loss combines a sigmoid activation and binary cross-entropy in one function. For each task, the loss is computed as:

$$\text{loss}(\hat{y}_k, y_k) = -\left(y_k \cdot \log(\sigma(\hat{y}_k)) + (1 - y_k) \cdot \log(1 - \sigma(\hat{y}_k))\right) \quad (6)$$

where $y_k$ is a one-hot representation of actual/gold label for the task $k$. $\sigma(\cdot)$ is the sigmoid function. To aggregate the individual losses for each task into a single weighted loss, we use the following formula:

$$\mathcal{L} = \sum_k w_k \frac{\sum \text{loss}(\hat{y}_k, y_k)}{N_k} \quad (7)$$

where $N_k$ is the number of classes of $k$ and $w_k$ is a learnable weight of $k$.

### 2.3 Fine-tuning Large Language Models for Relation Extraction

Large Language Models (LLMs) have demonstrated remarkable success across various biomedical NLP tasks due to their substantial growth and advancements. Recent studies (Chen, et al., 2024) indicate that Llama 3 (AI@Meta, 2024) exhibits performance comparable to GPT-3.5 (OpenAi, 2023) on several benchmarks. Leveraging parameter-efficient fine-tuning (PEFT) algorithms, such as QLORA (Dettmers, et al., 2024), enables the fine-tuning of Llama 3 with reduced computational resources. This study compares our approach against several state-of-the-art LLMs (Llama 3.2, and GPT-3.5 models) fine-tuned on biomedical data.

For the BioRED task, each prompt consists of four components: (1) role, which is always "You are a bioinformatics expert. "; (2) question, stated as "Identify the BioRED relation, direction, and novelty labels for the highlighted pair, '<<NE1 text>>' and '<<NE2 text>>,' and respond in JSON format," where "<<NE1 text>>" and "<<NE2 text>>" are the first mentioned named entities (NEs) in the text; (3) content, with <NE_type> .. </NE_type> tags used to mark the entity pair; and (4) response JSON, formatted as {"BioRED_relation_label": "Association", "direction_label": "Leftward", "novelty_label": "No"}. An example is shown in Fig 3.

```
{ "role":
            "You are a bioinformatics expert.",
    "prompt":
            "Identify the BioRED relation, direction, and novelty labels for the
            highlighted pair, \"type II diabetes\" and \"glucose\", and respond in JSON
            format.",
    "content":
            "... We therefore tested the hypothesis that variability in the HNF-6 gene is
            associated with subsets of <DiseaseOrPhenotypicFeature>Type II (non-insulin-
            dependent) diabetes mellitus</DiseaseOrPhenotypicFeature> and estimates of insulin
            secretion in <ChemicalEntity>glucose</ChemicalEntity> tolerant subjects. ...",
    "response":
            "{
                \"BioRED_relation_label\":\"Association\",
                \"direction_label\":\"Leftward\",
                \"novelty_label\":\"No\"
            }"}
```

## 3 Results

### 3.1 Datasets

Our model's performance was assessed on the two separate document-level RE benchmarks, BioRED and BC5CDR. Table 1 presents a statistical comparison of the latest BioRED corpus (Islamaj, et al., 2023) and BC5CDR (Wei, et al., 2016). In the latest BioRED, the training and development sets from the original BioRED 2022 corpus (Luo, et al., 2022) were combined and reused as the training set, while the test set from the original BioRED 2022 corpus was used as the development set. In addition, the 400 test abstracts used in the BioCreative VIII were used to evaluate and compare different methods. Early stopping is applied in our experiment based on the F1-score observed on the development set, where all hyperparameters are also optimized.

**Table 1.** Statistics of the enriched BioRED-BC8 and BC5CDR datasets.

| Statistics | BioRED-BC8 | BC5CDR |
|---|---|---|
| # training articles | 500 | 500 |
| # deve articles | 100 | 500 |
| # test articles | 400 | 500 |
| # NE pairs | 8 | 1 |
| # relation types | 8 | 1 |
| # relation/novelty | 12,536 | 3,116 |
| # directionality | 10,864 | - |

### 3.2 Evaluation metrics

For the BioRED task, we use the same evaluation metric as in the BioCreative VIII challenge, which employs the F1-measure to evaluate performance across four metrics. The four metrics for evaluation are:
- Entity pair (EP): Evaluate how predicted entity pairs match the gold standard.
- Entity pair + novelty label (EP+NV): Evaluate how predicted entity pairs and novelty labels match the gold standard.
- Relation type (RT): Evaluate how both predicted entity pairs and their relation types match the gold standard.
- Relation type + novelty label (RT+NV): Evaluate how predicted entity pairs, relation types, and novelty labels match the gold standard.

In the evaluation, "None" label is not counted as True Positive (TP) in F1-measure. In addition, we use the F1-measure to evaluate the performance of directionality determination as below:
- Relation type + directionality (RT+DI): Evaluate how predicted entity pairs, relation types, and directionalities match the gold standard.
- Relation type + directionality + novelty label (All): Evaluate how predicted entity pairs, relation types, directionalities, and novelty labels match the gold standard.

For the BC5CDR task, we applied the official document-level evaluation metric, which is based on the F1-measure and commonly used to assess systems on the BC5CDR dataset.

### 3.3 Experimental settings

We utilized Nvidia V100 and A100 GPUs for our experiments. Batch sizes of 4 and 8 were run on V100 GPUs, while a batch size of 16 required A100 GPUs for fine-tuning the model. All fine-tuning of LLama was performed using A100 GPUs, and ChatGPT models were fine-tuned with OpenAI playground. For the BC5CDR and BioRED datasets, we trained our models on the training set and optimized hyperparameters using the development set. We explored a range of hyperparameters, including learning rates (1e-5, 5e-6, and 1e-6), epoch counts (5, 10, and 20), batch sizes (4, 8, and 16), and soft prompt lengths (0, 8, 16, 24, and 32). The experiments involving PubMedBERT (Luo, et al., 2022), BioREx (Lai, et al., 2023), and our method were conducted ten times using different random seeds to enable statistically comparison. The results were averaged for stability. To report the final performance on BC5CDR, we retrained the model using the combined training and development sets with the optimal hyperparameters. For the BioRED, we did not retrain the model on the combined sets. The pre-trained model of BioREx was selected as the pre-trained language model due to its specialized biomedical knowledge, which we anticipated would enhance task performance.

Our experiments compared our method with state-of-the-art (SOTA) methods on the BioRED and BC5CDR tasks. In addition, we included recent LLMs, GPT, and Llama for comparison. For these models, we set the temperature to 0, the maximum sequence length to 1024, the batch size to 4, the number of epochs to 5, learning rate to 1e-4, and used 4-bit QLora quantization (Dettmers, et al., 2024). Note that the benchmark zero-shot results for GPT-3.5 and GPT-4 are sourced from previous reports (Islamaj, et al., 2024) and do not include directionality prediction. We did not evaluate GPT-4 with fine-tuning due to the associated cost considerations.

**Table 2.** Experiment results on the BioCreative VIII BioRED test set, reported as F1-scores (%). The evaluation includes entity pairs (EP), relation types (RT), directionality (DI), and novelty (NV) labels. '*' indicates statistically significant than PubMedBERT, and '**' indicates statistically significant than both PubMedBERT and BioREx

| Method | EP | EP+NV | RT | RT+NV | RT+DI. | All |
|---|---|---|---|---|---|---|
| GPT-3.5 (175B) + zero-shot (Islamaj, et al., 2024) | 37.44 | 22.83 | 13.41 | 8.90 | - | - |
| GPT-4 (1.76T) + zero-shot (Islamaj, et al., 2024) | 42.93 | 30.40 | 20.85 | 15.44 | - | - |
| LLama3.2-11B + zero-shot | 29.35 | 14.65 | 10.12 | 5.33 | 5.17 | 2.17 |
| GPT-3.5 (175B) + fine-tune | 70.18 | 53.84 | 50.01 | 38.52 | 42.81 | 33.22 |
| LLama3.2-11B + fine-tune | 68.78 | 54.36 | 47.78 | 37.54 | 40.80 | 32.39 |
| PubMedBERT (Luo, et al., 2022) | 74.07 | 56.19 | 52.69 | 39.93 | - | - |
| BioREx (Lai, et al., 2023) | 74.75* | 56.69 | 55.74* | 41.35* | - | - |
| Ours | **75.34**** | **57.26*** | **56.06**** | **42.66**** | **48.62** | **37.02** |

**Table 3.** Experiment results on the BC5CDR test set are reported as F1-scores (%). Δ refers to the standard derivation value.

| Method | Precision | Recall | F1 | ΔF1 |
|---|---|---|---|---|
| Llama 3.2-11B + zero-shot | 38.1 | 51.0 | 43.6 | - |
| GPT-3.5 (175B) + fine-tune | 64.3 | 76.8 | 70.0 | - |
| Llama 3.2-11B + fine-tune | 66.9 | 64.5 | 65.7 | - |
| EoG (Christopoulou, et al., 2019) | 62.1 | 65.6 | 63.6 | - |
| BERT-GT (Lai and Lu, 2021) | 64.9 | 67.1 | 66.0 | - |
| ATLOP (Zhou, et al., 2021) | 67.0 | 73.1 | 69.9 | 1.1 |
| HTG (Yuan, et al., 2024) | 62.4 | **78.1** | 69.4 | - |
| BioREx (Lai, et al., 2023) | 71.1 | 70.2 | 70.6 | 0.8 |
| Ours | **72.0** | 71.0 | **71.4** | 1.0 |

### 3.4 Results on the BioRED task

Table 2 presents the F1-scores of different models on the test set of the BioRED task, providing an overview of its performance across various metrics. As shown in Table 2, our method (Ours) demonstrates superior performance across most (four different) evaluation metrics, indicating its robust and versatile capabilities in biomedical relation extraction tasks. Ours achieves the highest performance on the overall metric. Moreover, the data-centric approach used by BioREx to combine nine different RE datasets cannot be extended to tasks such as novelty prediction. To address this, we integrated BioREx's relation predictions with the PubMedBERT-based novelty classification.

Due to the limited availability of comparable systems on the newer BioRED dataset, we compare our results with large language models (LLMs) and present results from GPT and fine-tuned versions of Llama3.2-11B and ChatGPT. Ours significantly outperforms these recent LLMs. Interestingly, our analysis shows that the fine-tuned LLama3.2-11B model performs comparably to PubMedBERT in the Novelty metrics despite PubMedBERT's domain-specific pre-training. This surprising result suggests that general-purpose language models, when fine-tuned using generative approaches (i.e., calculating loss through next-word prediction), can potentially rival the performance of domain-specific models in biomedical relation extraction tasks. These findings offer valuable insights for future research and model development, showing that no clear winner emerged among the evaluated LLMs. Nevertheless, ours consistently outperforms most configurations across all metrics.

When incorporating the additional complexity of relation directionality, our performance decreases. This decline in performance can be attributed to the increased complexity of the task, which we will discuss error cases in the discussion section. Despite this expected challenge, our method demonstrates robust performance compared to LLMs in the directionality determination task, highlighting the effectiveness of the multi-task learning approach.

### 3.5 Results on the BC5CDR dataset

Table 3 shows the F1-scores of ours on the BC5CDR test set. To ensure a fair comparison, we only include models that do not use external knowledge bases. We excluded systems that employed semantic segmentation, as we found they relied on a preprocessed BC5CDR dataset (Christopoulou, et al., 2019) where entities with chemical-induced disease (CID) relationships were ordered in a specific way. The ordering does not affect the EoG (Christopoulou, et al., 2019) approach; however, it can influence the generation of the entity-pair matrix for semantic segmentation, potentially impacting model performance. HTGRS reported a configuration without semantic segmentation (HTG), which is included in our comparison. As shown in Table 3, our method outperforms the compared models; however, the improvement over BioREx is not statistically significant. Nevertheless, the results demonstrate the robustness of our approach. Interestingly, GPT-3.5, with fine-tuning, outperforms most state-of-the-art models except ours. Interestingly, GPT-3.5, with fine-tuning, outperforms most state-of-the-art models except ours.

## 4 Discussion

### 5.1 Ablation Study

Table 4 presents the impact of different context chunks (prefix, suffix, and infix) on two test sets. We conducted individual evaluations for each chunk type to isolate the impact of each context chunk module. Contrary to our expectations, the infix chunk-only model is not the best performance among the three chunk types, especially on BC5CDR. The minimum window size containing two entities might result in unstable input sequence lengths, thus worsening the learning. However, each chunk model outperformed the other two single-chunk models in different metrics. Ours simultaneously leverages the complementary strengths of all three chunk types, resulting in a significant performance boost compared to any single-chunk model.

### 5.2 Effect of Soft Prompt Length

We investigated the impact of varying soft prompt lengths on model performance. Our findings, illustrated in Table 5, clearly demonstrate the efficacy of soft prompting. Notably, prompt length of 8 token yielded higher F1-scores than without soft prompt (where soft prompt length is 0) on both BioRED-BC8 and BC5CDR test sets, respectively. Importantly, a direct comparison between ours with and without soft prompting reveals that the inclusion of soft prompts leads to a marked improvement in performance across most evaluation metrics. Although the improvements did not pass the statistical test besides the EP metric, we believe that it still shows the potential of applying soft prompting as a technique to enhance the RE classification of PLM-based approaches.

Table 4. Experiment results of the ablation evaluation on the test sets of BioRED-BC8 and BC5CDR.

| Method | BioRED-BC8 | | | | | | BC5CDR |
|---|---|---|---|---|---|---|---|
| | Entity Pair | EP+Novelty | Relation Type | RT+Novelty | RT+Directionality | All | F1 |
| Ours (prefix chunk-only) | 74.38 | 55.27 | 53.86 | 40.03 | 44.88 | 32.43 | 70.3 |
| Ours (infix chunk-only) | 75.22 | 54.72 | 52.54 | 38.03 | 44.02 | 30.66 | 61.2 |
| Ours (suffix chunk-only) | 74.11 | 54.51 | 54.03 | 39.44 | 44.44 | 32.25 | 71.2 |
| Ours | 75.34 | **57.26** | 56.06 | **42.66** | **48.62** | **37.02** | 71.4 |

Table 5. Experiment results of the soft-prompt evaluation on the test sets of BioRED-BC8 and BC5CDR. '*' indicates statistically improvement.

| Method | BioRED-BC8 | | | | | | BC5CDR |
|---|---|---|---|---|---|---|---|
| | Entity Pair | EP+Novelty | Relation Type | RT+Novelty | RT+Directionality | All | F1 |
| Ours w/o soft-prompt | 74.54 | 56.79 | 55.50 | 42.23 | 47.78 | 36.46 | 71.0 |
| Ours | 75.34* | **57.26** | 56.06 | **42.66** | **48.62** | **37.02** | 71.4 |

### 5.3 Error Analysis and Limitation

The most common error type in chemical-chemical pairs involves False Negatives (FNs) and False Positives (FPs) in identifying the Cotreatment relation. In the test set, there are 172 Cotreatment pairs. However, approximately 48% of these are pure FNs, and ~49% are misclassified as either Negative_Correlation or Association. This issue arises due to the limited number of Cotreatment samples in the training set, which contains only 55 such pairs. We believe that data augmentation targeting specific relation types could help address this imbalance and reduce these errors. For chemical-disease pairs, the majority of errors involve indirect relationships that are implicitly inferred across multiple sentences. Approximately 74% of FN/FP cases in the test set are cross-sentence relationships. Regarding disease-variant relationships, error cases are often due to misclassification between Association and Positive_Correlation. For example, Positive_Correlation may be misclassified as Association and vice versa. Upon examining specific examples, we observed that misclassification often stems from the inherent ambiguity of natural language. For instance, in PMID:31298765, the abstract states: *"Sentence 0: SSBP1 mutations in dominant optic atrophy with variable retinal degeneration…. Sentence 6: RESULTS: We defined a new ADOA locus on 7q33-q35 and identified 3 different missense variants in SSBP1 (NM_001256510.1; c.113G>A [p.(Arg38Gln)], c.320G>A [p.(Arg107Gln)], and c.422G>A [p.(Ser141Asn)]) in affected individuals from 2 families and 2 singletons with ADOA and variable retinal degeneration."* In this case, the variants (Sentence 6) have a "Positive_Correlation" relationship with "optic atrophy," as stated in the title (Sentence 0 "SSBP1 mutations"). However, those variants are "Association" with "retinal degeneration" rather than "Positive_Correlation."

Furthermore, based on our analysis of error cases involving directionality, the most challenging cases are gene-gene and chemical-gene relationships. When directionality was taken into consideration, their performance dropped by 15% and 11%, respectively. These challenges primarily stem from the prevalence of "Association" and "Bind" relation types, which are inherently difficult to classify with a specific directionality due to the complexity of natural language. For example, in PMID:33491092, the sentence states: *"TNFalpha/IL-17A-induced IL-36gamma expression also involved the nuclear factor kappaB (NF-kappaB), p38 mitogen-activated protein kinase, and ERK1/2 signalling pathways."* The ground truth identifies "IL-17A" as the subject and "NF-kappaB" as the object of the Association relationship. However, the biomedical mechanisms between them are not explicit within this context.

Finally, in this work, we proposed a new model for the BioRED, BC5CDR, and directionality tasks, leveraging the capabilities of deep learning models and LLMs. However, the biomedical domain offers a wealth of external resources and knowledge that hold significant potential for enhancing the performance of our method. Similarly, recent advancements (Xiong, et al., 2024) in retrieval-augmented generation (RAG) research demonstrate its potential to enhance the capabilities of LLMs. We look forward to exploring these resources further to address the current performance bottlenecks.

## 5 Conclusion

The rapid expansion of biomedical literature poses a significant challenge in accurately extracting critical relationships between biological entities. To address this, we enriched the BioRED dataset by manually annotating 1,000 abstracts for directionality, encompassing 10,864 directional relationships. This comprehensive annotation facilitates a deeper understanding of causal interactions, which is crucial for downstream applications like drug development and personalized medicine. Our method leverages multi-task learning to handle three BioRED tasks (relation pair, novelty, and directionality) simultaneously. Additionally, ours employs context chunking to manage long contexts and further enhances the performance through soft prompt learning. Our method demonstrates robust performance in predicting directionality on the BioRED task, significantly surpassing its predecessor, BioREx, particularly on the BC5CDR test set. These advancements in biomedical relation extraction improve the efficiency and accuracy of BioRED and establish a new benchmark for future developments in the field.

## Acknowledgements

We would like to thank Donald Comeau for his invaluable assistance in proofreading this manuscript and providing thoughtful feedback, which greatly improved the clarity and quality of the paper. We are grateful to Qiao Jin for his help in running the fine-tuning of ChatGPT.

## Funding

This research was supported by the NIH Intramural Research Program, National Library of Medicine.

*Conflict of Interest:* none declared.